\documentclass[11pt]{article}
\usepackage{amsmath}
\usepackage{amsfonts}
\usepackage{amssymb}
\usepackage{booktabs}
\usepackage{longtable}
\usepackage{multirow}
\usepackage{rotating}
\usepackage{color}
\usepackage{graphics}
\usepackage{graphicx}
\usepackage{epsfig} 
\usepackage{subfigure}
\usepackage{latexsym,bm}
\usepackage{hhline}
\usepackage[greek,english]{babel}
\usepackage{mathrsfs}
\usepackage{makecell}
\usepackage{threeparttable}
\usepackage[ruled,linesnumbered]{algorithm2e}
\usepackage[strings]{underscore}
\usepackage{authblk}
\usepackage{setspace}

\usepackage[]{hyperref}
\hypersetup{
     colorlinks=true,
     linkcolor=blue,
     filecolor=gray,
     urlcolor=blue,
     citecolor=blue,
 }

\newtheorem{theorem}{Theorem}

\newtheorem{definition}[theorem]{Definition}
\newtheorem{example}{Example}

\begin{document}

\title{PMNN: Physical Model-driven Neural Network for solving time‑fractional differential equations}

\author[1]{Zhiying Ma}
\author[2]{Jie Hou}
\author[2]{Wenhao Zhu}
\author[1*]{Yaxin Peng}
\author[2*]{Ying Li}
\affil[1]{Department of Mathematics, School of Science, Shanghai University, Shanghai, 200444, PR China}
\affil[2]{School of Computer Engineering and Science, Shanghai University, Shanghai, 200444, PR China}

\date{}
\maketitle
\let\thefootnote\relax\footnotetext{*Corresponding author: yinglotus@shu.edu.cn (Ying Li), yaxin.peng@shu.edu.cn (Yaxin Peng)}

\begin{abstract}
\noindent
In this paper, an innovative Physical Model-driven Neural Network (PMNN) method is proposed to solve time‑fractional differential equations. It establishes a temporal iteration scheme based on physical model-driven neural networks which effectively combines deep neural networks (DNNs) with interpolation approximation of fractional derivatives. Specifically, once the fractional differential operator is discretized, DNNs are employed as a bridge to integrate interpolation approximation techniques with differential equations. On the basis of this integration, we construct a neural-based iteration scheme. Subsequently, by training DNNs to learn this temporal iteration scheme, approximate solutions to the differential equations can be obtained. The proposed method aims to preserve the intrinsic physical information within the equations as far as possible. It fully utilizes the powerful fitting capability of neural networks while maintaining the efficiency of the difference schemes for fractional differential equations. Moreover, we validate the efficiency and accuracy of PMNN through several numerical experiments.
\end{abstract}

\noindent{Keywords: Deep neural network; Physical model-driven; Iteration scheme approximation; Time-fractional differential equation}
\section{Introduction}
In the study of fractional differential equations (FDEs), researchers have observed that fractional-order differential operators possess non-local properties, which distinguishes them from integer-order differential operators. As a result, FDEs are well-suited for describing dynamic processes in the real world that involve memory and hereditary characteristics. FDEs have been widely applied in various fields \cite{book1} such as anomalous diffusion, viscoelasticity, fluid mechanics, electromagnetic waves, statistical models, signal processing and system identification, quantum economics, fractal theory, robotics, etc. However, it is extremely challenging to obtain analytical solutions for FDEs. Even if exact solutions can be obtained, they may involve complex functions like Mittag-Leffler functions, H functions, Wright functions, and so on \cite{function}. Dealing with these functions in numerical computations is complicated. Therefore, finding effective numerical simulation methods for fractional differential equations has become one of the important research topics. In recent decades, researchers have made significant advancements in the numerical solution of FDEs. Various numerical solution techniques have been developed, including finite difference method (FDM) \cite{book1, FDM1}, finite element method (FEM) \cite{FEM1, FEM2}, spectral methods \cite{spectral1, spectral2, spectral3}, Wavelet methods \cite{Wavelet1, Wavelet2, Wavelet3}, matrix methods \cite{matrix1, matrix2, matrix3}, Laplace transforms \cite{Laplace1}, variational iteration methods \cite{variational1}, and Adomian decomposition methods \cite{Adomi1, Adomi2, Adomi3}. These traditional numerical methods have made significant advancements in the solving of fractional-order  equations and dealing with fractional order derivatives. These advancements have laid the foundation for other innovative numerical methods of FDEs.

With the rapid advancement of deep learning technology, an increasing number of scholars have embarked on the exploration of employing deep learning for solving differential equations. As one of fundamental models in the field of deep learning, DNNs exhibit not only remarkable fitting capabilities but also the capacity to learn and optimize models in an adaptive manner. Consequently, the utilization of deep learning for solving differential equations has emerged as a burgeoning research direction that gains significant attention. Currently, a multitude of neural network-based methods have been proposed for solving integer-order differential equations. For instance, Lagaris et al. were among the pioneers who successfully applied Artificial Neural Networks (ANNs) to solve integer-order differential equations. They employed ANNs to construct trial solutions for solving initial value and boundary value problems \cite{trail}. Raissi et al. introduced the Physics-Informed Neural Networks (PINNs) approach, a new class of universal function approximators that is capable of encoding any underlying physical laws present in a given dataset \cite{PINN}. Due to its high predictive accuracy and robustness, PINNs rapidly became a benchmark method in the field, leading many researchers to conduct related studies \cite{gPINN,APINN,PPINN,nPINN,canPINN}. Lu et al. introduced DeepONets, a deep operator network, to accurately and effectively learn nonlinear operators from relatively small datasets \cite{DeepONet}. These approaches mentioned above are all based on data-driven deep neural network methods,  which are characterized by their fast prediction speed. The features required by data-driven methods depend only on the training data \cite{2022survey}. In practical applications, it is important to note that differential equations contain many physical laws. Therefore, relying solely on sampled data to capture information may result in incomplete findings. For certain complex equations, even if enough training data is selected, it remains challenging to accurately capture the physical information encoded within the equations. This poses a disadvantage to solving differential equations.

In fact, among the various neural network methods for solving integer-order differential equations, there is a class of methods known as model-driven approaches. These methods do not require a large number of sample data for their construction. For example, Li et al. proposed an iterative scheme approximation based on deep learning framework, known as DeLISA. The authors first obtained time iteration schemes using implicit multistep and Runge-Kutta methods. Subsequently, this iteration scheme was approximated using a neural network. This method achieves continuous-time prediction without the need for a large number of interior points \cite{delisa}. Long et al. utilized Convolutional Neural Networks (CNNs) to develop a novel approach for solving non-stationary partial differential equations, known as PDE-Net. The fundamental idea of this method is to employ convolutional kernels to learn differential operators and use neural networks to approximate unknown nonlinear responses \cite{PDENet}. Building upon their earlier PDE-Net framework, Long et al. proposed a new deep neural network called PDE-Net 2.0. It combines numerical approximation of differential operators using convolutions with multi-layer symbolic neural networks for model recovery, which is employed to discover potential differential equations from observed data \cite{PDENet2}. These model-driven methods possess clear statistical or physical significance, and many traditional methods can be directly combined with them, which provides a new perspective for solving differential equations.

However, deep neural networks face a major challenge when it comes to solving fractional  differential equations: the handling of fractional derivatives. This is due to the limitations of automatic differentiation techniques, which can only compute derivatives of integer orders.  Consequently, deep neural networks are unable to handle fractional derivatives  directly.  As a result, research on utilizing neural networks for solving FDEs is relatively limited, but several novel methods have been proposed by scholars. 

Raja et al. successfully solved various types of linear and nonlinear FDEs using feedforward ANNs. Unlike other numerical methods, this approach offers the advantage of providing a continuous solution over the entire finite domain \cite{nn-ode}. 
Zúñiga-Aguilar et al. proposed a new ANN method to approximate the solution of FDEs. The author mainly considered the variable-order fractional differential equations with Mittag-Leffler kernel in the sense of Liouville-Caputo. \cite{2017CSF}. 
Pang et al. extended PINNs to fractional PINNs (fPINNs) for solving spatiotemporal fractional advection-diffusion equations (ADEs), and systematically studied their convergence. The authors utilized automatic differentiation to analytically compute integer-order derivatives within equations. While for fractional-order derivatives, they utilize traditional numerical approximation methods for numerical discretization. This approach effectively overcomes the challenge of neural networks being unable to directly solve fractional derivatives \cite{fPINN}. 
Qu et al. developed neural networks based on sine and cosine functions using uniformly distributed sampling points. They obtained approximate solutions to initial boundary value problems of several FDEs \cite{sin-cos,sin-cos1}. 
Ye et al. employed Physics-Informed Neural Networks (PINNs) to investigate the forward and inverse problems of time fractional diffusion equations with conformable derivative, addressing the limitation of directly applying neural networks to solve fractional-order differential equations \cite{pinn-fde}. 
Fang et al. solved the fractional PDEs in high-dimension and the inverse problems using DNNs.\cite{20D}. 
Therefore, it naturally raises the question of whether the combination of traditional numerical methods and model-driven neural network methods can be utilized to solve FDEs. The answer is affirmative.

Motivated by the aforementioned work, we propose an innovative Physical Model-driven Neural Network (PMNN) method for solving FDEs. In PMNN, we first discretize the fractional derivatives using an interpolation-based Finite Difference Method (FDM) to construct a temporal iteration scheme for the equation. Subsequently, we use this iteration scheme to construct the loss function for training a physical model-driven neural network. Since the iteration scheme contains the physical information of the equation, obtaining an approximate solution can be viewed as the learning process of the temporal iteration scheme using a physical model-driven neural network. Therefore, when the loss gradually decreases and converges, we can consider the neural network as an approximation solution of the equation. The proposed method preserves the physical information of the equation to the greatest extent. Furthermore, it effectively utilizes the powerful fitting capability of neural networks while ensuring the validity of the difference schemes for fractional differential equations. Moreover, in this paper, we evaluate the performance of PMNN through several numerical experiments. These experiments demonstrate the effectiveness of the method and its potential in solving a wide range of fractional differential equation problems. In addition, it is worth noting that the PMNN method, introduced in this paper, represents a general framework for solving FDEs. It combines neural networks with fractional derivative interpolation approximations, without being limited to a specific approximation method. In this paper, we employ two interpolation approximation methods: $\mathrm{L}1$ and $\mathrm{L} 2-1_\sigma$, which result in two different physical model-driven neural network: PMNN on $\mathrm{L}1$ and PMNN on $\mathrm{L} 2-1_\sigma$. 

The remaining part of this paper is organized as follows: In Section 2, we first introduce the fundamental knowledge of FDEs, and then provide a comprehensive overview of the $\mathrm{L}1$ and $\mathrm{L} 2-1_\sigma$ interpolation approximation for fractional derivatives. In Section 3, we present a detailed description of the construction of PMNN and provide a step-by-step explanation of the process involved in solving equations using PMNN. Section 4 primarily presents several numerical experiments to demonstrate the effectiveness of our proposed method. Finally, Section 5 provides a summary of the paper.

\section{Preliminaries}
\label{2}
In this section, we will review two definitions of fractional derivatives and two interpolation approximations for the Caputo derivatives.
\subsection{Definitions}
We start by presenting several fundamental definitions, which can be found in \cite{li2015numerical}.
\begin{definition}
    The Riemann-Liouville integral with order $\alpha>0$ of the given function $f(t)$, $t\in(a, b)$ are defined as
	\begin{equation}
	\begin{aligned}
	{ }_{a}\mathrm{I}_{t}^{-\alpha} f(t)=\frac{1}{\Gamma(\alpha)} \int_a^t(t-s)^{\alpha-1} f(s) \mathrm{d}s,	
	\end{aligned}
	\end{equation}
	where $\Gamma(\cdot)$ is the Euler's gamma function.
\end{definition}

\begin{definition}
	The Riemann-Liouville derivatives with order $\alpha>0$ of the given function $f(t)$, $t\in(a, b)$ are defined as
	\begin{equation}
	\begin{aligned}
	{ }_{a} \mathrm{D}_{t}^\alpha f(t) & =\frac{\mathrm{d}^m}{\mathrm{~d} t^m}\left[{ }_{a}\mathrm{I}_{t}^{-(m-\alpha)} f(t)\right] \\
    & =\frac{1}{\Gamma(m-\alpha)} \frac{\mathrm{d}^m}{\mathrm{~d} t^m} \int_a^t(t-s)^{m-\alpha-1} f(s) \mathrm{d} s,
	\end{aligned}
	\end{equation}
	where $m$ is a positive integer satisfying $m-1 \leq \alpha<m$.

\end{definition}

\begin{definition}
	The Caputo derivatives with order $\alpha>0$ of the given function $f(t)$, $t \in(a, b)$ are defined as	
	\begin{equation}
	\begin{aligned}
	{ }_a^C \mathrm{D}_{t}^\alpha f(t) & ={ }_{a}\mathrm{I}_{t}^{-(m-\alpha)}\left[f^{(m)}(t)\right] \\
    & =\frac{1}{\Gamma(m-\alpha)} \int_a^t(t-s)^{m-\alpha-1} f^{(m)}(s) \mathrm{d} s	,
    \end{aligned}
	\end{equation}
	where $m$ is a positive integer satisfying $m-1<\alpha \leq m$.
\end{definition}

The Riemann-Liouville (R-L) derivative and the Caputo derivative may exhibit differences in numerical computations. For instance, the Caputo fractional derivative of a constant function is 0, whereas the R-L fractional derivative is non-zero. The distinct characteristics of these two integrals determine their applicability in different contexts. The R-L derivative imposes fewer conditions on the function $f(x)$, enhancing its convenience for mathematical theoretical investigations. On the other hand, the Caputo derivative finds wider application in solving initial and boundary value problems of differential equations in the field of physical engineering. In this paper, we adopt the Caputo derivative.
\subsection{Interpolation approximation of Caputo derivative}
Over the past several decades, researchers in the field of Finite Difference Methods for solving FDEs have made significant research achievements. The underlying idea of these methods is to discretize the fractional derivatives in the differential equation, transforming the fractional-order equations into integer-order equations for numerical computation. The interpolation approximation of the fractional derivative serves as a crucial step in the FDMs, which provides us with a promising direction for solving FDEs using DNNs. Following that, we will illustrate two interpolation approximations for the $\alpha$-order Caputo fractional derivative.

\subsubsection{$\mathrm{L}1$ approximation}
\label{section-l1}
For the Caputo derivative of order $\alpha$ $(0<\alpha<1)$
\begin{equation}
\begin{aligned}
{ }_0^C D_t^\alpha f(t)=\frac{1}{\Gamma(1-\alpha)} \int_0^t \frac{f^{\prime}(s)}{(t-s)^\alpha} \mathrm{d} s,
\end{aligned}
\end{equation}
the most commonly used approach is the $\mathrm{L}1$ approximation based on piecewise linear interpolation.

Let $N$ be a positive integer. We define $\tau=\frac{T}{N}$, $t_k=k \tau$, $0 \leqslant k \leqslant N$, and
\begin{equation}
\begin{aligned}
a_l^{(\alpha)}=(l+1)^{1-\alpha}-l^{1-\alpha}, \quad l \geqslant 0,
\end{aligned}
\end{equation}
\noindent 
we derive an approximation formula for the calculation of ${ }_0^C D_t^\alpha f(t)|_{t=t_n}$:
\begin{equation}
\begin{aligned}
\label{L1}
D_t^\alpha f(t_n) \equiv \frac{\tau^{-\alpha}}{\Gamma(2-\alpha)}\left[a_0^{(\alpha)} f(t_n)-\sum_{k=1}^{n-1}\left(a_{n-k-1}^{(\alpha)}-a_{n-k}^{(\alpha)}\right) f(t_k)-a_{n-1}^{(\alpha)} f(t_0)\right].
\end{aligned}
\end{equation}
Eq.\eqref{L1} is commonly known as the $\mathrm{L}1$ formula or $\mathrm{L}1$ approximation.

\subsubsection{$\mathrm{L} 2-1_\sigma$ approximation}
\label{section-l2-1}
For the $\alpha$ $(0<\alpha<1)$ order Caputo derivative, the aforementioned $\mathrm{L}1$ approximation formula achieves uniform convergence of order $2-\alpha$. Alikhanov \cite{L2-1sigma} further extended this result by discovering superconvergent interpolation points and establishing the $\mathrm{L} 2-1_\sigma$ approximation formula, which achieves uniform convergence of order $3-\alpha$. In the following section, we will present this result in detail. Let us denote
\begin{equation}
\begin{aligned}
\sigma=1-\frac{\alpha}{2}, \quad t_{n+\sigma}=(n+\sigma) \tau, \quad f^n=f(t_n),
 \end{aligned}
\end{equation}
the approximation formula for evaluating ${ }_0^C D_t^\alpha f(t)|_{t=t_{n-1+\sigma}}$ can be obtained as follows:
\begin{equation}
\begin{aligned}
\label{L2-1}
\Delta_t^\alpha f(t_{n-1+\sigma}) \equiv \frac{\tau^{-\alpha}}{\Gamma(2-\alpha)} \sum_{k=0}^{n-1} c_k^{(n, \alpha)}[f(t_{n-k})-f(t_{n-k-1})], \quad 1 \leqslant n \leqslant N .
\end{aligned}
\end{equation}
Eq.\eqref{L2-1} is typically referred to as the $\mathrm{L} 2-1_\sigma$ formula or $\mathrm{L} 2-1_\sigma$ approximation. When $n=1$,
\begin{equation}
\begin{aligned}
c_0^{(1, \alpha)}=\sigma^{1-\alpha},
\end{aligned}
\end{equation}
while when $n \geqslant 2$,
\begin{equation}
\left\{\begin{aligned}
c_0^{(n, \alpha)} = & \frac{(1+\sigma)^{2-\alpha}-\sigma^{2-\alpha}}{2-\alpha}-\frac{(1+\sigma)^{1-\alpha}-\sigma^{1-\alpha}}{2}, \\
c_k^{(n, \alpha)} = & \frac{1}{2-\alpha}[(k+1+\sigma)^{2-\alpha}-2(k+\sigma)^{2-\alpha}+(k-1+\sigma)^{2-\alpha}] \\
& -\frac{1}{2}[(k+1+\sigma)^{1-\alpha}-2(k+\sigma)^{1-\alpha}+(k-1+\sigma)^{1-\alpha}], \\
& 1 \leqslant k \leqslant n-2, \\
c_{n-1}^{(n, \alpha)} = & \frac{1}{2} [3(n-1+\sigma)^{1-\alpha}-(n-2+\sigma)^{1-\alpha}] \\
&-\frac{1}{2-\alpha}[(n-1+\sigma)^{2-\alpha}-(n-2+\sigma)^{2-\alpha}] .
\end{aligned}\right.
\end{equation}

\subsection{Classification of Caputo Fractional Partial Differential Equations}
Consider the following Caputo fractional partial differential equation:
\begin{equation}
\begin{aligned}
{ }_0^C D_t^\alpha u=\Delta u
\end{aligned}
\end{equation}
Based on the interval of values for $\alpha$, it can be categorized \cite{li2019theory}, as depicted in Table \ref{T-eq classify}. This paper primarily addresses the case involving derivatives of order $0<\alpha<1$.

\begin{table}[!h]
\caption{The classification of ${ }_0^C D_t^\alpha u=\Delta u$ with $\alpha\in (0, 2]$.}\label{T-eq classify}
 \begin{center}
  \begin{tabular}[]{ccc} 
  \toprule[1pt]
  $\alpha$ & Math.type &  Phys.sense \\
  \hline
  $(0,1)$ & Time-fractional parabolic equation & Temporal subdiffusion \\
  1 & Parabolic equation & Diffusion\\
  $(1,2)$ & Time-fractional hyperbolic equation & Temporal Supperdiffusion \\
  2 & Hyperbolic equation & Wave \\
  \bottomrule[1pt]
  \end{tabular}
 \end{center}
\end{table}

\section{Illustration of the method}

\subsection{Problem setup}
In this paper, we focus on fractional ordinary differential equations (FODEs) and fractional partial differential equations (FPDEs). Next, we provide a detailed exposition of our methodology by considering the initial-boundary value problem for the FPDE on a bounded domain $\Omega\in \mathbb{R}^{n}$.

Consider the initial-boundary value problem for the following time-fractional slow diffusion equation:
\begin{align}
{ }_0^C D_t^\alpha u(\boldsymbol{x},t) & = \mathcal{L}u(\boldsymbol{x},t)+f(\boldsymbol{x},t), \quad (\boldsymbol{x},t) \in \Omega \times (0, T], \label{eq of 1D PDE}\\
u(\boldsymbol{x},t) & =\mu(\boldsymbol{x},t), ~~~~~~~~~~~~~~\quad (\boldsymbol{x},t) \in \partial\Omega \times (0, T],\label{bound of 1D PDE}\\
u(\boldsymbol{x},0) & =\varphi(\boldsymbol{x}), ~~~~~~~~~~~~~~~~~\quad \boldsymbol{x} \in \Omega, \label{int of 1D PDE}
\end{align}
where, ${}_0^C D_t^\alpha$ denotes the Caputo fractional derivative with $\alpha \in (0,1)$, $\mathcal{L}$ represents an integer-order differential operator, $u(\boldsymbol{x},t)$ is the solution of the equation, $f$, $\mu$ and $\varphi$ are known functions.

\subsection{Architecture of PMNN}
DNNs have gained remarkable achievements in tackling differential equations of integer order, owing largely to the integration of automatic differentiation techniques. Nevertheless, the applicability of automatic differentiation is restricted to functions with integer-order differentials. To surmount the obstacle of automatic differentiation in utilizing neural networks for solving fractional-order differentials, we tackle the issue by discretizing the Caputo derivatives in the equations using $\mathrm{L}1$ and $\mathrm{L} 2-1_\sigma$ approximations, respectively. Subsequently, neural networks are introduced to establish PMNN on $\mathrm{L}1$ and PMNN on $\mathrm{L} 2-1_\sigma$, respectively. This section presents the detailed procedures involved in constructing PMNN.

\subsubsection{PMNN on $\mathrm{L}1$}
The first step of our approach is to semi-discretize the fractional-order derivative in the temporal domain. Let $N$ be a positive integer. We define $\tau=\frac{T}{N}$, $t_k=k\tau$, where $0 \leqslant k \leqslant N$. Utilizing the $\mathrm{L}1$ formula presented in Section \ref{section-l1}, we obtain:
\begin{equation}
\begin{aligned}
{ }_0^C D_t^\alpha u^{n} &\approx D_t^\alpha u_{n} \\
&= \frac{\tau^{-\alpha}}{\Gamma(2-\alpha)}\left[a_0^{(\alpha)} u^{n}-\sum_{k=1}^{n-1}\left(a_{n-k-1}^{(\alpha)}-a_{n-k}^{(\alpha)}\right) u^{k}-a_{n-1}^{(\alpha)} u^{0}\right],
\end{aligned}
\end{equation}
where $u_{n} = u(t_n, \boldsymbol{x})$, $1 \leqslant n \leqslant N$. Substituting it into the governing equation, we obtain the temporal iteration scheme based on Eq.\eqref{eq of 1D PDE}:
\begin{equation}
\begin{aligned}
u^{n} = \frac{\Gamma(2-\alpha)\cdot\tau^{\alpha}}{a_0^{(\alpha)}}\left[ \mathcal{L}u^{n} + f^{n} \right] + \sum_{k=1}^{n-1}\left(\frac{a_{n-k-1}^{(\alpha)}-a_{n-k}^{(\alpha)}}{a_0^{(\alpha)}}\right) u^{k} + \frac{a_{n-1}^{(\alpha)}}{a_0^{(\alpha)}}u^{0}.
\label{l1 1d}
\end{aligned}
\end{equation}

\begin{figure}[!ht]
\centering
\includegraphics[width=6.2in]{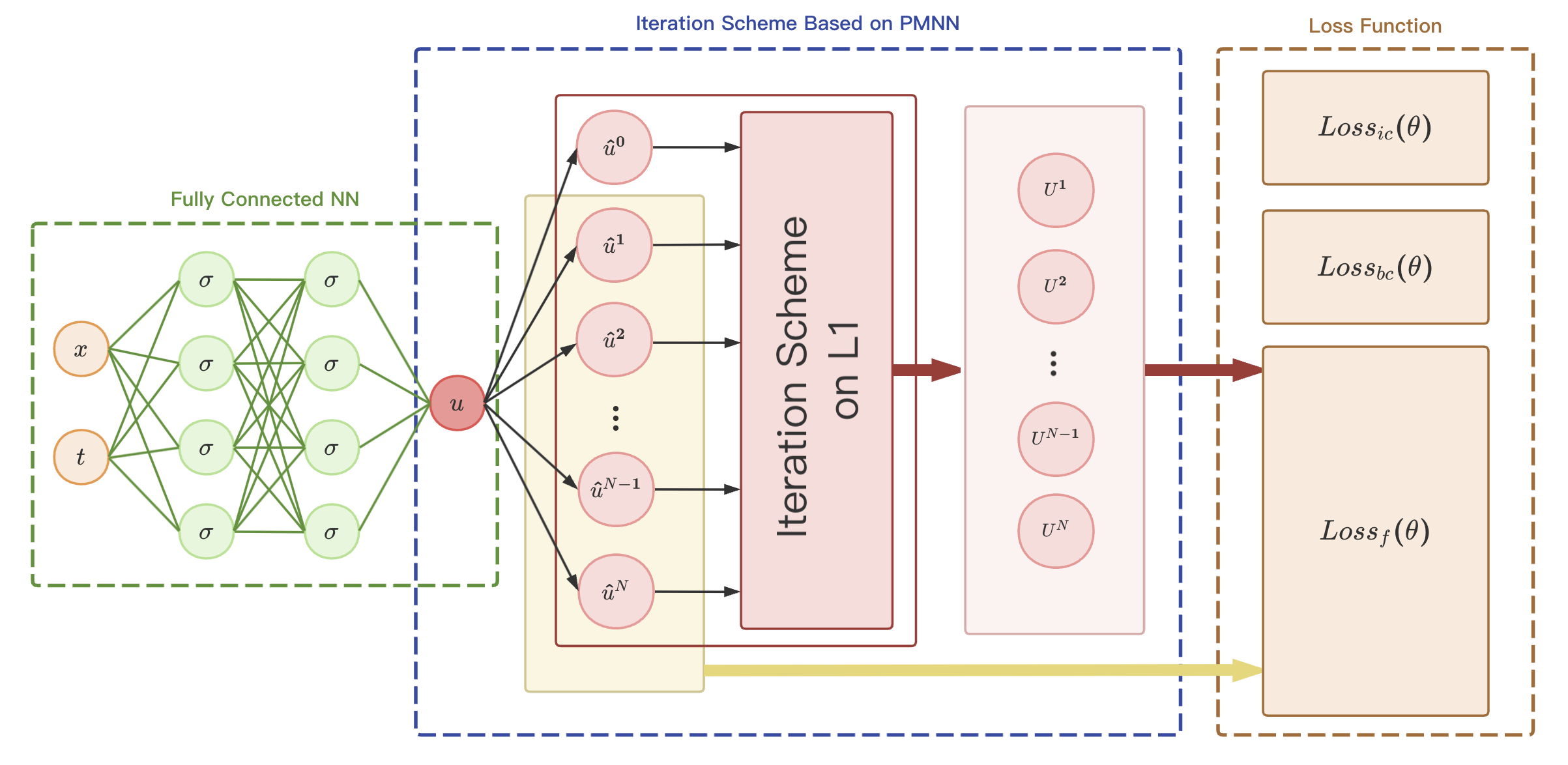}
\caption{The architecture of PMNN on $\mathrm{L}1$} 
\label{NN1}
\end{figure}

Next, we introduce physical model-driven neural networks as solvers to obtain approximate solutions for the differential equations. In this study, we directly consider the output of the neural network, $\hat{u}(\boldsymbol{x}, t;\theta)$, as the approximation solution. By substituting it into Eq.\eqref{l1 1d}, we obtain the expression for PMNN on $\mathrm{L}1$ as follows:
\begin{equation}
\begin{aligned}
U^{n} = \frac{\Gamma(2-\alpha)\cdot\tau^{\alpha}}{a_0^{(\alpha)}}\left[ \mathcal{L}\hat{u}^{n} + f^{n} \right] + \sum_{k=1}^{n-1}\left(\frac{a_{n-k-1}^{(\alpha)}-a_{n-k}^{(\alpha)}}{a_0^{(\alpha)}}\right) \hat{u}^{k} + \frac{a_{n-1}^{(\alpha)}}{a_0^{(\alpha)}}\hat{u}^{0},
\end{aligned}
\end{equation}
where $\hat{u}^{n}=u(\boldsymbol{x}, t_{n};\theta)$ represents the output of the neural network at time  $t_{n}$, and $f^{n}=f(\boldsymbol{x}, t_{n})$. Evidently, the iterative scheme mentioned above incorporates the physical information inherent in the governing equation. Taking into account the initial and boundary conditions \eqref{bound of 1D PDE}-\eqref{int of 1D PDE}, we define the loss function for the PMNN on $\mathrm{L}1$ model as follows:
\begin{equation}
\begin{aligned}
Loss(\theta)= Loss_{f}(\theta)+Loss_{ic}(\theta)+Loss_{bc}(\theta),
\end{aligned}
\end{equation}
the definitions of each component in the loss function are as follows:
\begin{equation}
\begin{aligned}
Loss_{f}(\theta)&= \frac{1}{N_{f}}\sum_{i=1}^{N_{f}}\left[\hat{u}(x^{i}_{f}, t^{i}_{f}) - U(x^{i}_{f}, t^{i}_{f})\right]^{2},\\
Loss_{bc}(\theta)&= \frac{1}{N_{bc}}\sum_{i=1}^{N_{bc}}\left[\hat{u}(x^{i}_{bc}, t^{i}_{bc}) - u(x^{i}_{bc}, t^{i}_{bc})\right]^{2},\\
Loss_{ic}(\theta)&= \frac{1}{N_{ic}}\sum_{i=1}^{N_{ic}}\left[\hat{u}(x^{i}_{ic}, t^{i}_{bc}) - u(x^{i}_{ic}, t^{i}_{ic})\right]^{2},
\label{loss3}
\end{aligned}
\end{equation}
where $\{x^{i}_{f}, t^{i}_{f}\}_{i=1}^{N_{f}}$ represents the training points of the iterative scheme, where $N_{f}$ denotes the number of training points. $\{x^{i}_{ic}, t^{i}_{ic}\}_{i=1}^{N_{ic}}$ refers to the initial training points, where $N_{ic}$ represents the number of initial training points. Similarly, $\{x^{i}_{bc}, t^{i}_{bc}\}_{i=1}^{N_{bc}}$ denotes the boundary points, and $N_{bc}$ represents the number of boundary points. The architecture of the PMNN on $\mathrm{L}1$ model is illustrated in Fig.\ref{NN1}.

\subsubsection{PMNN on $\mathrm{L} 2-1_\sigma$}
Just like in the case of PMNN on $\mathrm{L}1$, we proceed with the discretization of the equation. Let $N$ be a positive integer. We define $\tau=\frac{T}{N}$, $t_k=k \tau$, $0 \leqslant k \leqslant N$, and $\sigma=1-\frac{\alpha}{2}$. Based on the $\mathrm{L} 2-1_\sigma$ formula described in Section \ref{section-l2-1}, we obtain:
\begin{equation}
\begin{aligned}
{ }_0^C D_t^\alpha u^{n-1+\sigma} &\approx \Delta_t^\alpha u(t_{n-1+\sigma}) \\
&=  \frac{\tau^{-\alpha}}{\Gamma(2-\alpha)} \sum_{k=0}^{n-1} c_k^{(n, \alpha)}[u(t_{n-k})-u(t_{n-k-1})],
\end{aligned}
\end{equation}
where $u_{n} = u(t_n, \boldsymbol{x})$, $1 \leqslant n \leqslant N$. We note that the $\mathrm{L} 2-1_\sigma$ scheme, unlike the $\mathrm{L}1$ scheme, discretizes the fractional derivative at time points $t_{n-1+\sigma}$, which is located outside the set of discrete time points $t_n$. Therefore, in the process of discretizing the equation, we need to consider the time node $t_{n-1+\sigma}$. The temporal iteration scheme, derived from the differential equation (\ref{eq of 1D PDE}), is given by:
\begin{equation}
\begin{aligned}
u^{n} = \frac{\Gamma(2-\alpha)\cdot\tau^{\alpha}}{c_0^{(n, \alpha)}}\left[\mathcal{L}u^{n-1+\sigma} + f^{n-1+\sigma}\right] + \sum_{k=1}^{n-1} \frac{c_k^{(n, \alpha)}}{c_0^{(n, \alpha)}}(u^{n-k-1}-u^{n-k}) + u^{n-1}. \label{l2-1 1d}
\end{aligned}
\end{equation}

\begin{figure}[!ht]
\centering
\includegraphics[width=6.2in]{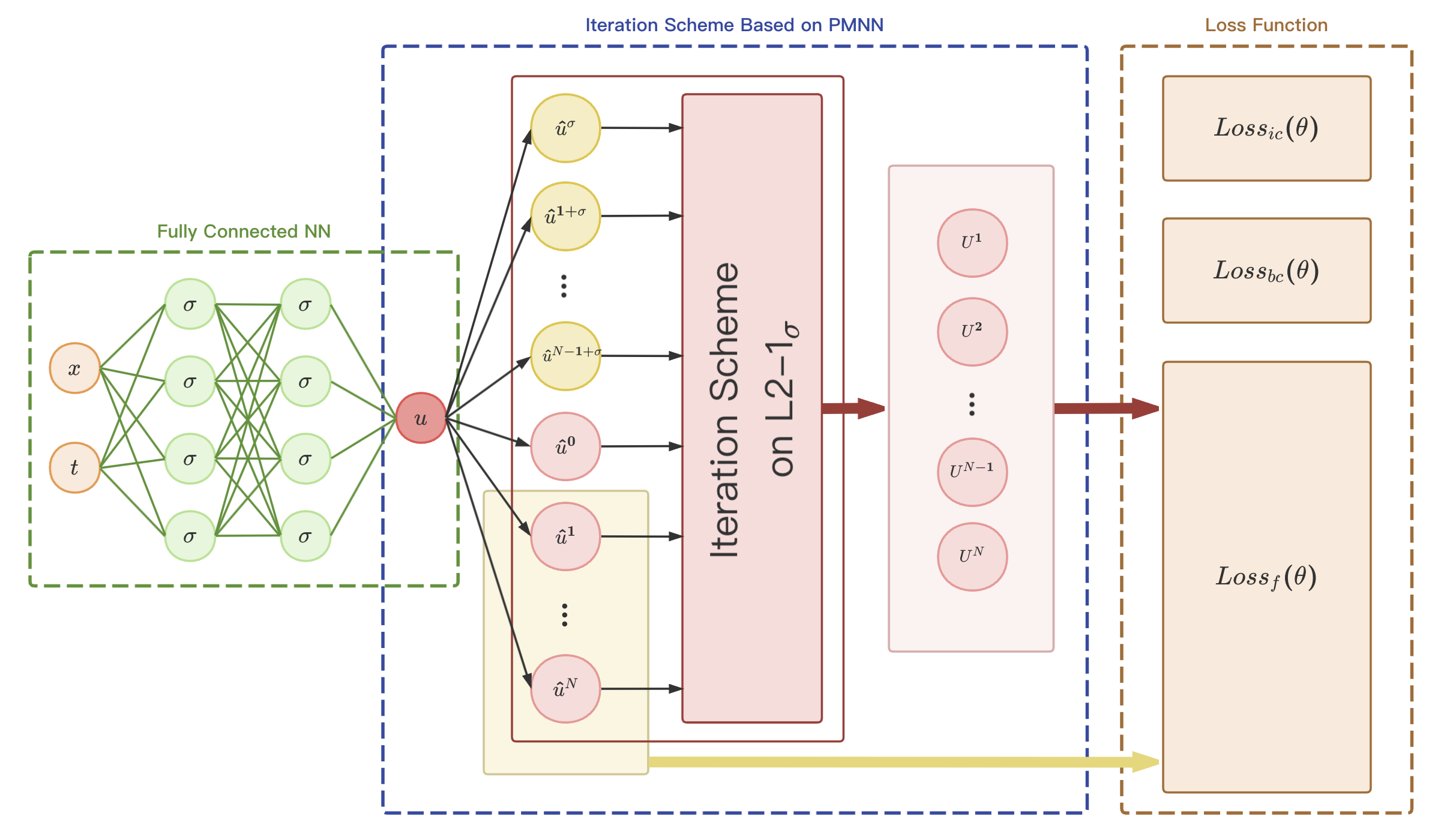}
\caption{The architecture of PMNN on $\mathrm{L} 2-1_\sigma$} 
\label{NN2}
\end{figure}

Similarly, by substituting the neural network's output $\hat{u}(\boldsymbol{x}, t;\theta)$ as an approximation into equation (\ref{l2-1 1d}), we can derive the expression for PMNN on $\mathrm{L} 2-1_\sigma$ as follows:
\begin{equation}
\begin{aligned}
U^{n} = \frac{\Gamma(2-\alpha)\cdot\tau^{\alpha}}{c_0^{(n, \alpha)}}\left[\mathcal{L}\hat{u}^{n-1+\sigma} + f^{n-1+\sigma}\right] + \sum_{k=1}^{n-1} \frac{c_k^{(n, \alpha)}}{c_0^{(n, \alpha)}}(\hat{u}^{n-k-1}-\hat{u}^{n-k}) + \hat{u}^{n-1},
\end{aligned}
\end{equation}
where $\hat{u}^{n}=u(\boldsymbol{x}, t_{n};\theta)$ represents the output of the network at time $t_{n}$, and $f^{n}=f(\boldsymbol{x}, t_{n})$. The loss function for the PMNN on $\mathrm{L} 2-1_\sigma$ model is defined by
\begin{equation}
\begin{aligned}
Loss(\theta)= Loss_{f}(\theta)+Loss_{ic}(\theta)+Loss_{bc}(\theta).
\end{aligned}
\end{equation}
The definitions of each loss term can be found in Eq.\eqref{loss3}. The architecture of the PMNN on $\mathrm{L} 2-1_\sigma$ model is depicted in Fig.\ref{NN2}.

%
%
%
%

\section{Numerical results}
\label{3}
In this section, the performance of our proposed model is evaluated through the analysis of three single-term temporal fractional differential equations: a time-fractional ODE, a one-dimensional time-fractional PDE, and a two-dimensional time-fractional PDE. The loss function is minimized using the L-BFGS-B method, and the accuracy of the numerical solutions is assessed using the $L^{2}$ relative error. By contrasting the effectiveness of the two models, it is observed that the proposed model achieves high accuracy. Additionally, all experiments in this section are carried out utilizing an NVIDIA RTX 1660 GPU card.

\begin{example}\label{example1} \rm{
Single-term Temporal Fractional Ordinary Differential Equation

Considering a single-term temporal fractional ordinary differential equation:
\begin{equation}
\begin{aligned}
{ }_0^C D_t^\alpha u(t)&=-u(t) + f(t), \quad t\in(0,T], \\
u(0)&=0,\\
\end{aligned}
\end{equation}
where, $0<\alpha<1$, and the exact solution to the equation is given by $u(t)=t^{5+\alpha}$. The right-hand side term is defined as $f(t)= \frac{\Gamma(6+\alpha)}{120}t^{5}+t^{5+\alpha}$. For this experiment, we focus on the case where $T=1$.

In this experiment, a fully connected neural network (FNN) with 5 hidden layers is utilized. Each hidden layer consists of 20 neurons with a tanh activation function. The training and testing data are uniformly sampled from the interval $[0, T]$. For our analysis, we specifically select $N_t$ training points and 500 testing points. To begin, we conduct tests for the case of $\alpha=0.5$ and $N_t=41$. Fig.\ref{ODE_pred} shows the comparison between the predicted solutions of the two models and the exact solution. The blue solid line represents the graph of the exact solution, while the red dashed line represents the curve of the predicted solution using the PMNN model. It is evident that the predicted solution precisely aligns with the exact solution.

\begin{figure}[!h]
\centering
\subfigure[PMNN on $L1$]{
\includegraphics[width=2.5in]{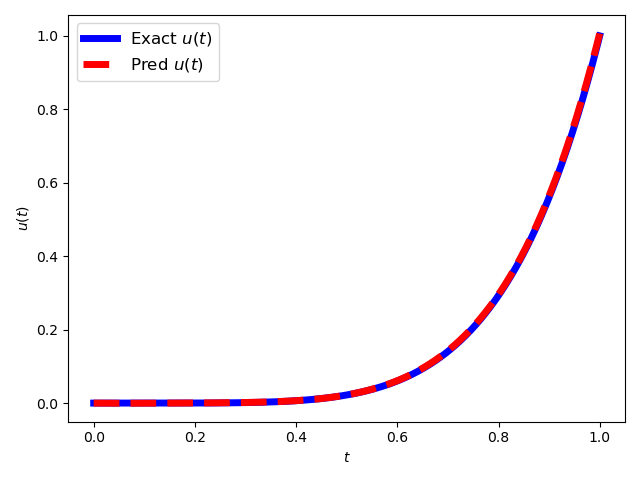}}
\subfigure[PMNN on $\mathrm{L} 2-1_\sigma$]{
\includegraphics[width=2.5in]{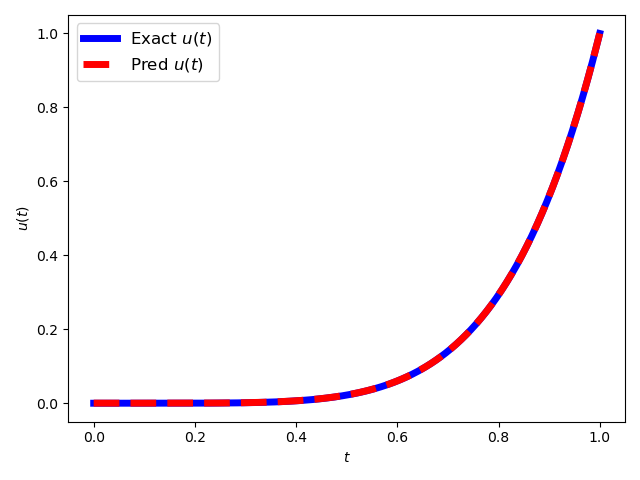}}
\caption{Single-term FODE: the exact solution and the predict solution of PMNN.} 
\label{ODE_pred}
\end{figure}

Fig.\ref{ODE_error} illustrates the variation of the error with respect to time $t$. For the PMNN on $\mathrm{L}1$ model, the error initially remains below $1 \times 10^{-3}$, but around $t=0.4$, it gradually increases over time.  In contrast, the error for the PMNN on $\mathrm{L} 2-1_\sigma$ model remains stable at around $1 \times 10^{-4}$. Therefore, in this case, the PMNN on $\mathrm{L} 2-1_\sigma$ model demonstrates superior overall performance.
\begin{figure}[!h]
\centering
\subfigure[PMNN on $L1$]{
\includegraphics[width=2.5in]{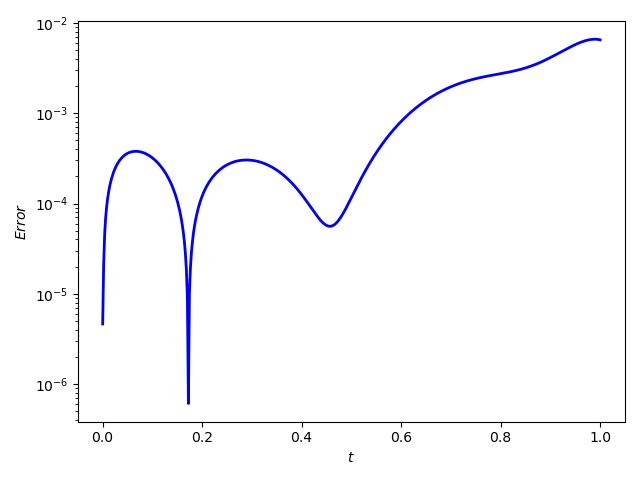}}
\subfigure[PMNN on $\mathrm{L} 2-1_\sigma$]{
\includegraphics[width=2.5in]{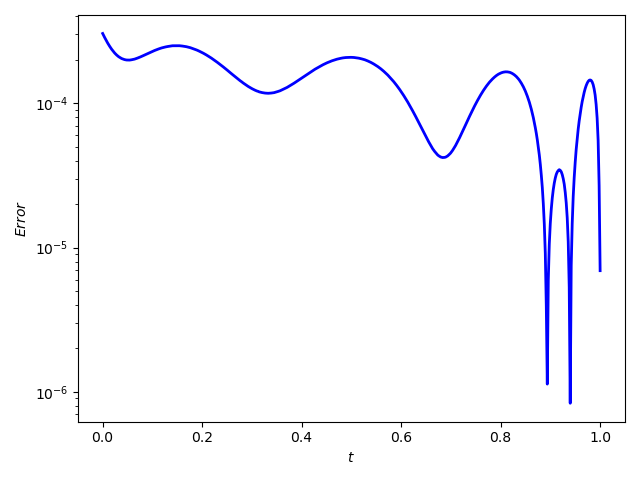}}
\caption{Single-term FODE: the trend of the error with the time $t$.} 
\label{ODE_error}
\end{figure}

The trends of loss with respect to the number of iterations for the two models are depicted in Fig.\ref{ODE_loss}. A comparison of the plots reveals that the PMNN on $\mathrm{L} 2-1_\sigma$ model exhibits faster convergence. Table \ref{tab:ODE_time} displays the iteration counts and training times for the two models with different $\alpha$. Comparing the iteration counts, it is evident that PMNN on $\mathrm{L} 2-1_\sigma$ converges more rapidly. However, it needs a longer training time, potentially due to its requirement for a larger training dataset.

\begin{figure}[!h]
\centering
\subfigure[PMNN on $L1$]{
\includegraphics[width=2.0in]{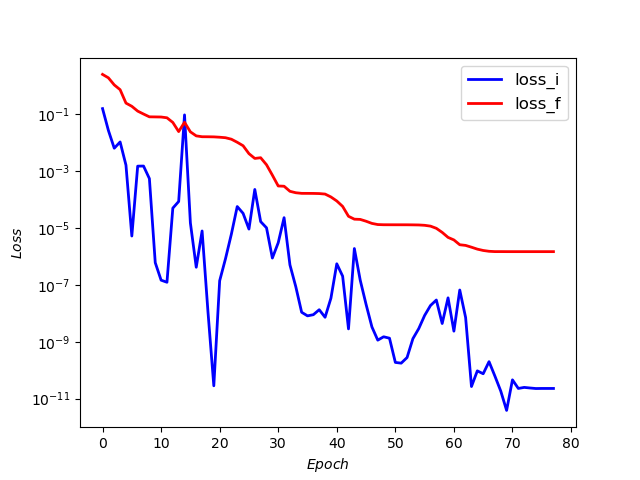}}
\subfigure[PMNN on $\mathrm{L} 2-1_\sigma$]{
\includegraphics[width=2.0in]{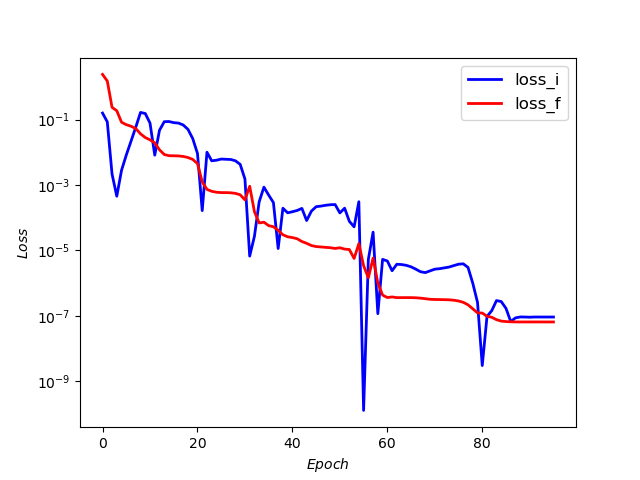}}
\subfigure[Comparison of the two models]{
\includegraphics[width=2.0in]{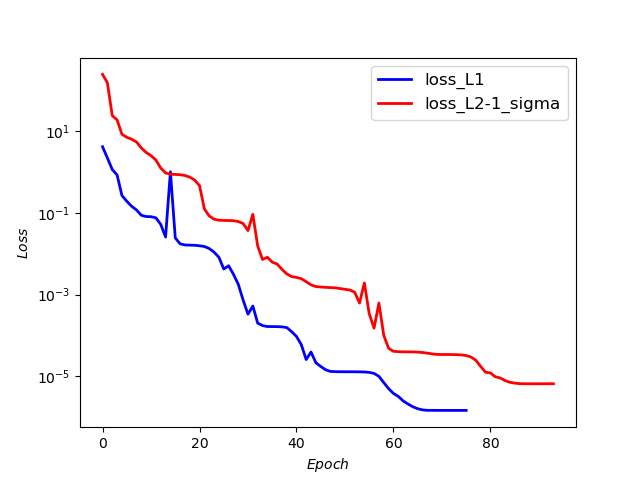}}
\caption{Single-term FODE: the trend of the loss function with the number of iterations} 
\label{ODE_loss}
\end{figure}

\begin{table}[!h] 
\caption{The number of iterations and training time of two PMNN for single-term FODE}
\label{tab:ODE_time}
\begin{center}
\begin{tabular}{c cc cc}
    \toprule
    \multirow{2}{*}{}& 
    \multicolumn{2}{c}{Iter} & \multicolumn{2}{c}{Training Time(s)}\cr
    \cmidrule(lr){2-3} \cmidrule(lr){4-5} 
    & $L1$ & $L2-1_\sigma$ & $L1$ & $L2-1_\sigma$ \cr
    \midrule
    $\alpha=0.25$ & 90 & 77 & 8.41 & 8.12  \cr
    $\alpha=0.50$ & 77 & 73 & 6.47 & 13.99 \cr
    $\alpha=0.75$ & 94 & 71 & 8.49 & 12.57  \cr
    \bottomrule
   \end{tabular}
\end{center}
\end{table}

To further evaluate the performance of the proposed method, we conducted tests with different $\alpha$ and varying numbers of training points. The $L^{2}$ relative errors for each case are presented in Table \ref{tab:ODE}. It can be observed that as the number of training points increases, the overall errors of both models decrease. Moreover, after approximately $N_{t}=41$, both models demonstrate good performance. The minimum error attained by PMNN on $\mathrm{L}1$ is $1.71 \times 10^{-4}$, whereas for PMNN on $\mathrm{L} 2-1_\sigma$, the minimum error reaches $3.52 \times 10^{-4}$.
\begin{table}[!h]
  \caption{The $L^{2}$ error of two PMNN for single-term FODE}
  \label{tab:ODE}
  \begin{center}
    \begin{tabular}{c cc cc cc}
    \toprule
    \multirow{2}{*}{$N_{t}$}& 
    \multicolumn{2}{c}{$\alpha=0.25$} & \multicolumn{2}{c}{$\alpha=0.50$} & \multicolumn{2}{c}{$\alpha=0.75$} \cr
    \cmidrule(lr){2-3} \cmidrule(lr){4-5} \cmidrule(lr){6-7}
    & $L1$ & $L2-1_\sigma$ & $L1$ & $L2-1_\sigma$ & $L1$ & $L2-1_\sigma$\cr
    \midrule
    11 & 1.55 e-02 & 1.20 e-03 & 5.31 e-02 & 2.50 e-03 & 1.33 e-01 & 6.28 e-03 \cr
    21 & 5.35 e-03 & 1.21 e-03 & 2.07 e-02 & 2.19 e-03 & 5.88 e-02 & 1.53 e-03 \cr
    41 & 1.86 e-03 & 3.71 e-03 & 7.84 e-03 & 3.02 e-03 & 2.55 e-02 & 7.84 e-04 \cr
    81 & 9.58 e-04 & 2.64 e-03 & 3.10 e-03 & 1.58 e-03 & 1.09 e-02 & 4.75 e-04 \cr
    101 & 6.68 e-04 & 3.51 e-03 & 2.26 e-03 & 2.51 e-03 & 8.36 e-03 & 3.39 e-04 \cr
    201 & 1.71 e-04 & 3.86 e-03 & 1.26 e-03 & 2.37 e-03 & 4.59 e-03 & 3.52 e-04 \cr
    \bottomrule
    \end{tabular}
    \end{center}
\end{table}

}
\end{example}

\begin{example}\label{example2} \rm{
1D Time-fractional Convection-diffusion Equation

Consider the following one-dimensional time-fractional convection-diffusion equation, with the exact solution given by $u(x,t)=x^{2}+\frac{2t^{\alpha}}{\Gamma(1+\alpha)}$.\\
\begin{equation*}
\begin{aligned}
\frac{\partial^{\alpha} u(x,t)}{\partial t^{\alpha}}&=u_{xx},~~~~~~~~~~~~~~~~~x\in[0, 1],~~t\in(0, 1],\\
u(x,0)&=x^{2},~~~~~~~~~~~~~~~~~~~x\in[0, 1],\\
u(0,t)&=\frac{2t^{\alpha}}{\Gamma(1+\alpha)},~~~~~~~~~~t\in[0, 1],\\
u(1,t)&=1+\frac{2t^{\alpha}}{\Gamma(1+\alpha)},~~~~~t\in[0, 1],
\end{aligned}
\end{equation*}

The same FNN architecture is employed in this experiment, following the configuration of Example \ref{example1}. We uniformly sample $N_{t}$ time nodes from the interval $[0,1]$ and $N_{x}$ spatial nodes from the interval $[0,1]$. Consequently, the training data comprises a total of $N_{t} \times N_{x}$ data points. Similarly, we select $100 \times 100$ test data points. We begin the experiment by fixing $\alpha=0.5$, $N_{t}=41$, and $N_{x}=11$. Fig.\ref{1D_PDE_pred} shows the comparison between the predicted solutions obtained from the two models and the exact solution. The left panel depicts the graph of the exact solution, while the middle and right panels display the predicted solutions using PMNN on $\mathrm{L}1$ and PMNN on $\mathrm{L} 2-1_\sigma$, respectively. A visual analysis reveals a remarkable consistency between the predicted solutions and the graph of the exact solution.

\begin{figure}[!h]
\centering
\subfigure[Exact]{
\includegraphics[width=2.0in]{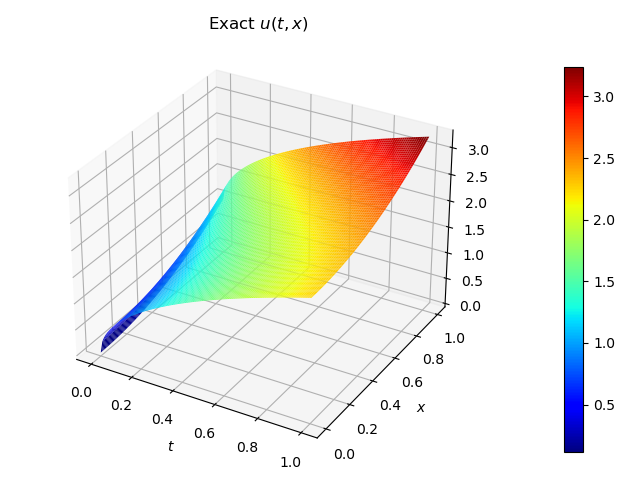}}
\subfigure[PMNN on $L1$]{
\includegraphics[width=2.0in]{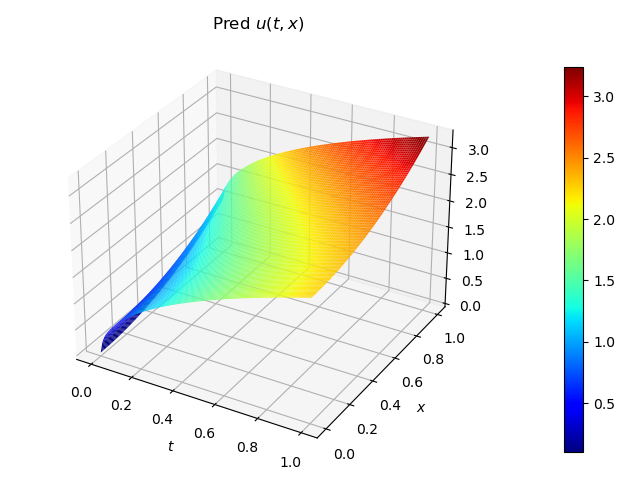}}
\subfigure[PMNN on $\mathrm{L} 2-1_\sigma$]{
\includegraphics[width=2.0in]{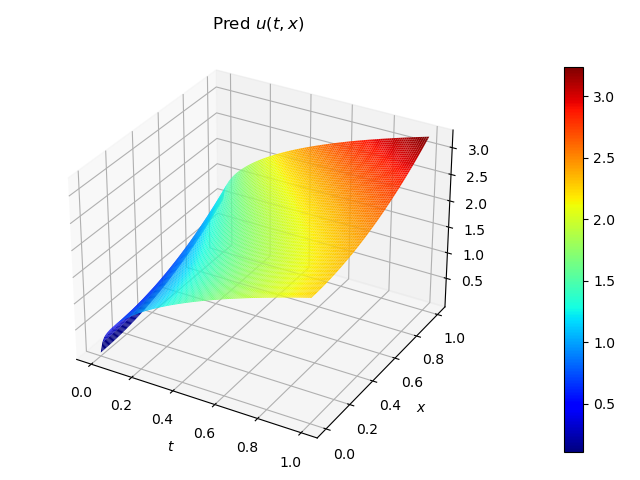}}
\caption{1D Single-term FPDE: the exact solution and the predict solution of PMNN.} 
\label{1D_PDE_pred}
\end{figure}

The error of the two PMNN models is depicted in Fig.\ref{1D_PDE_error_1}. It is evident that in this case, there are fluctuations in the error within a small interval near $t=0$, while the error remains stable for the rest of the time. Furthermore, it can be observed that during the fluctuation period, the error exhibits symmetry around the spatial coordinate point $x=0.5$. This is an intriguing phenomenon. Additionally, the error of PMNN on $\mathrm{L} 2-1_\sigma$ converges to a stable state in a shorter time frame.

\begin{figure}[!h]
\centering
\subfigure[PMNN on $L1$]{
\includegraphics[width=2.5in]{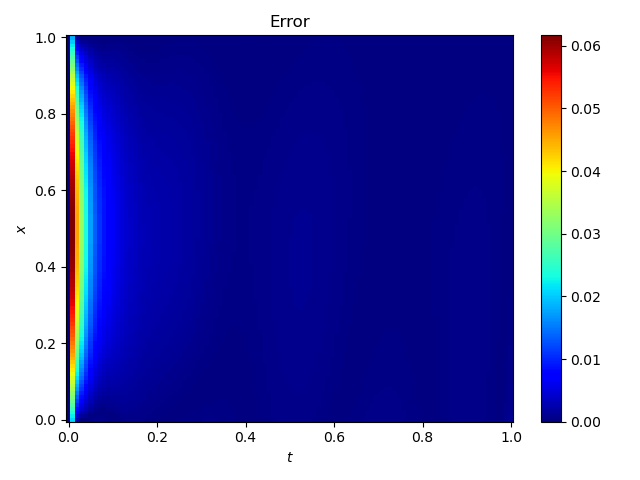}}
\subfigure[PMNN on $\mathrm{L} 2-1_\sigma$]{
\includegraphics[width=2.5in]{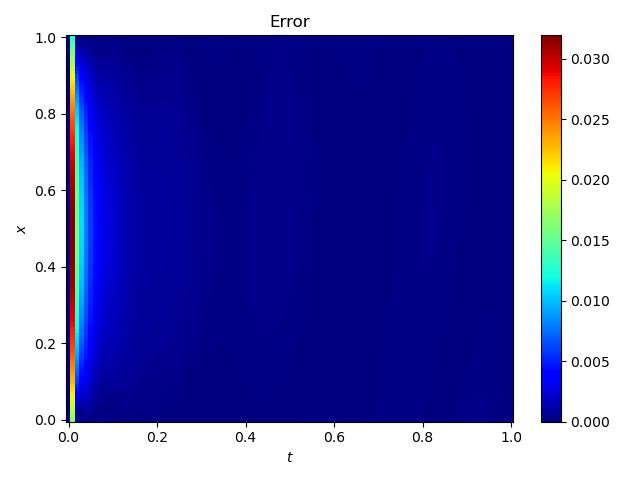}}
\caption{1D Single-term FPDE: the error presentation of PMNN.} 
\label{1D_PDE_error_1}
\end{figure}


The evolution of the loss functions for both models during the iterative process is visualized in Fig.\ref{1D_PDE_loss}. The first two subfigures in Fig.\ref{1D_PDE_loss} offer a comprehensive depiction of the dynamic evolution of the individual components comprising the loss functions, whereas the last subfigure portrays the overall variation of the loss. By comparing the two plots, it becomes apparent that PMNN on $\mathrm{L} 1$ demonstrates a noticeably faster convergence rate, which is contrary to the observations in Example \ref{example1}.
\begin{figure}[!h]
\centering
\subfigure[PMNN on $L1$]{
\includegraphics[width=2.0in]{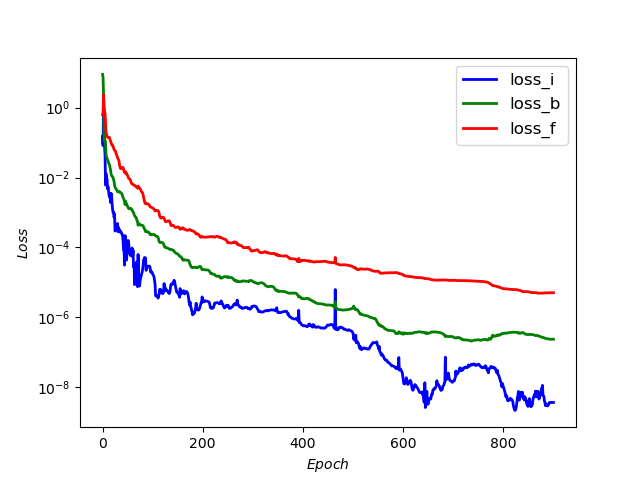}}
\subfigure[PMNN on $\mathrm{L} 2-1_\sigma$]{
\includegraphics[width=2.0in]{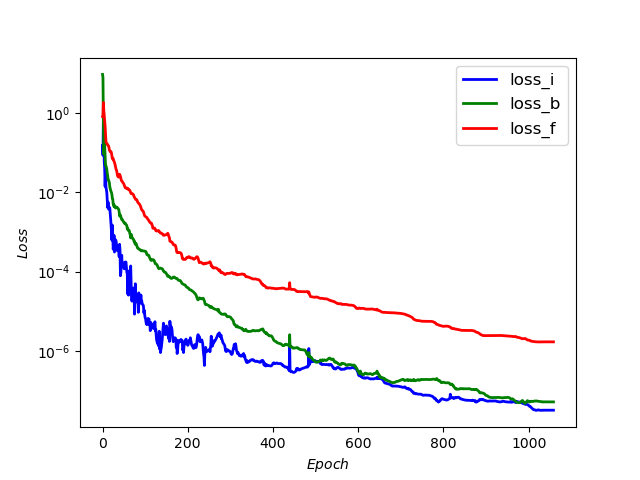}}
\subfigure[Comparison of the two models]{
\includegraphics[width=2.0in]{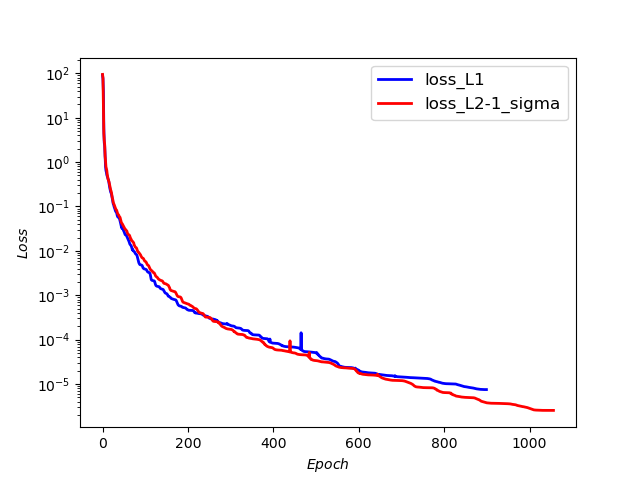}}
\caption{1D Single-term FPDE: the trend of the loss function with the number of iterations.} 
\label{1D_PDE_loss}
\end{figure}


\newpage
Table \ref{tab:1D_PDE_time} provides the iteration counts and training times for the two models with varying orders $\alpha$. It is observed that PMNN on $\mathrm{L}1$ converges faster when $\alpha=0.25$ and $\alpha=0.5$, whereas PMNN on $\mathrm{L}2-1_\sigma$ converges faster when $\alpha=0.75$. This indicates that the performance of the two models is influenced by the choice of order in this particular example.
\begin{table}[!h] 
  \caption{The number of iterations and training time of two PMNN for 1D single-term FPDE}
  \label{tab:1D_PDE_time}
  \begin{center}
    \begin{tabular}{c cc cc}
    \toprule
    \multirow{2}{*}{}& 
    \multicolumn{2}{c}{Iter} & \multicolumn{2}{c}{Training Time(s)}\cr
    \cmidrule(lr){2-3} \cmidrule(lr){4-5} 
    & $L1$ & $L2-1_\sigma$ & $L1$ & $L2-1_\sigma$ \cr
    \midrule
    $\alpha=0.25$ & 919 & 689 & 97.06 & 103.28	  \cr
    $\alpha=0.50$ & 901 & 1058 & 170.55 & 194.2  \cr
    $\alpha=0.75$ & 1630 & 1511 & 307.21 & 233.91  \cr
    \bottomrule
    \end{tabular}
   \end{center}
\end{table}

In the following experiments, we fix the number of time nodes at $N_{t}=41$ and set $\alpha=0.5$. We then change the number of spatial nodes, denoted as $N_{x}$, to investigate the influence on the models. The corresponding $L^{2}$ relative errors for each case are presented in Table \ref{tab:1D_PDE_x}. It can be observed that as the number of spatial training points increases, the errors of both models exhibit minimal fluctuations. This suggests that the quantity of spatial training points has minimal impact on the model's performance. Therefore, it is feasible to select a smaller number of spatial nodes for training, thereby reducing computational costs effectively.
\begin{table}[!h] 
  \caption{The $L^{2}$ error of two PMNN for 1D single-term FPDE}
  \label{tab:1D_PDE_x}
  \begin{center}
    \begin{tabular}{c cc}
    \toprule
    \multirow{2}{*}{$N_{x}$}& 
    \multicolumn{2}{c}{$\alpha=0.5$} \cr
    \cmidrule(lr){2-3}
    & $L1$ & $L2-1_\sigma$ \cr
    \midrule
    6   & 3.24e-03 & 1.75e-03 \cr
    11  & 3.66e-03 & 1.64e-03 \cr
    21  & 3.45e-03 & 1.59e-03 \cr
    41  & 3.37e-03 & 1.61e-03 \cr
    81  & 3.52e-03 & 1.68e-03 \cr
    101 & 3.59e-03 & 1.63e-03 \cr
    \bottomrule
    \end{tabular}
   \end{center}
\end{table}

Finally, we fix $N_{x}=11$ and perform tests with different values of $\alpha$ and varying numbers of spatial nodes $N_{x}$. The $L^{2}$ relative errors are presented in Table \ref{tab:1D_PDE}. Consistent with the experimental results in Example \ref{example1}, the errors of both models decrease as the number of time nodes increases. Overall, PMNN on $\mathrm{L} 2-1_\sigma$ outperforms PMNN on $\mathrm{L} 1$ in terms of performance.
\begin{table}[!h] 
  \caption{The $L^{2}$ error of two PMNN for 1D single-term FPDE}
  \label{tab:1D_PDE}
  \begin{center}
    \begin{tabular}{c cc cc cc}
    \toprule
    \multirow{2}{*}{$N_{t}$}& 
    \multicolumn{2}{c}{$\alpha=0.25$} & \multicolumn{2}{c}{$\alpha=0.50$} & \multicolumn{2}{c}{$\alpha=0.75$} \cr
    \cmidrule(lr){2-3} \cmidrule(lr){4-5} \cmidrule(lr){6-7}
    & $L1$ & $L2-1_\sigma$ & $L1$ & $L2-1_\sigma$ & $L1$ & $L2-1_\sigma$\cr
    \midrule
    11 & 3.46e-02 & 2.80e-02 & 1.38e-02 & 8.03e-03 & 6.39e-03 & 2.14e-03 \cr
    21 & 1.81e-02 & 1.58e-02 & 6.43e-03 & 3.59e-03 & 3.52e-03 & 1.17e-03 \cr
    41 & 7.74e-03 & 5.85e-03 & 3.66e-03 & 1.64e-03 & 2.03e-03 & 6.70e-04 \cr
    81 & 2.21e-03 & 9.73e-04 & 1.85e-03 & 6.74e-04 & 1.13e-03 & 3.81e-04 \cr
    101 & 1.43e-03 & 5.24e-04 & 1.48e-03 & 5.28e-04 & 9.61e-04 & 2.90e-04 \cr
    \bottomrule
    \end{tabular}
    \end{center}
\end{table}

}
\end{example}

\begin{example}\label{example3} \rm{
2D Time-fractional Convection-diffusion Equation

Consider the following time-fractional convection-diffusion equation, with an exact solution given by $u(\boldsymbol{x},t)=t^{2}e^{x+y}$.\\
\begin{equation*}
\begin{aligned}
\frac{\partial^{\alpha} u(\boldsymbol{x},t)}{\partial t^{\alpha}}&=\Delta u(\boldsymbol{x},t)+f(\boldsymbol{x},t),~~~~\boldsymbol{x}\in\Omega\subset \mathbb{R}^{2},~~t\in(0,1),\\
u(\boldsymbol{x},t)&=t^{2}e^{x+y},~~~~~~~~~~~~~~~~~~~\boldsymbol{x}\in\partial\Omega,~~t\in(0,1),\\
u(\boldsymbol{x},0)&=0,~~~~~~~~~~~~~~~~~~~~~~~~~~\boldsymbol{x}\in\Omega,\\
\end{aligned}
\end{equation*}
where $f(\boldsymbol{x},t)=[\frac{2t^{2-\alpha}}{\Gamma(3-\alpha)}-2t^{2}]e^{x+y}$. $\Omega=[0,1]\times[0,1]$

The present experiment follows the same configuration as the previous two experiments using FNN. We uniformly select $N_{t}$ time nodes on the interval $[0,1]$ and $N_{x}\times N_{x}$ spatial nodes on the domain $[0,1]\times [0,1]$, resulting in a total of $N_{t} \times N_{x} \times N_{x}$ training data points. In a similar manner, $100 \times 100 \times 100$ test data points are selected. For the experiment, we set $\alpha=0.5$, $N_{t}=21$, and $N_{x}=11$. Fig.\ref{2D_PDE_pred} illustrates the comparison between the exact solution and the predicted solutions obtained by the two models at $t=1$. On the left is the image of the exact solution, in the middle is the image of the predicted solution using PMNN on $\mathrm{L}1$, and on the right is the image of the predicted solution using PMNN on $\mathrm{L} 2-1_\sigma$. Through a visual comparison, it is evident that the predicted solution is the same as the exact solution.
\begin{figure}[!h]
\centering
\subfigure[Exact]{
\includegraphics[width=2.0in]{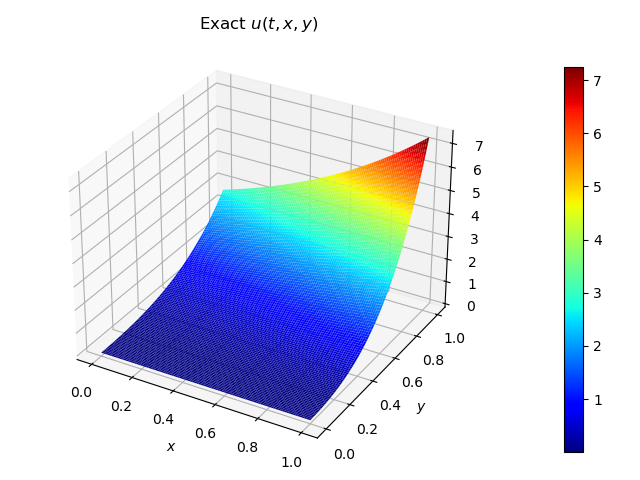}}
\subfigure[PMNN on $L1$]{
\includegraphics[width=2.0in]{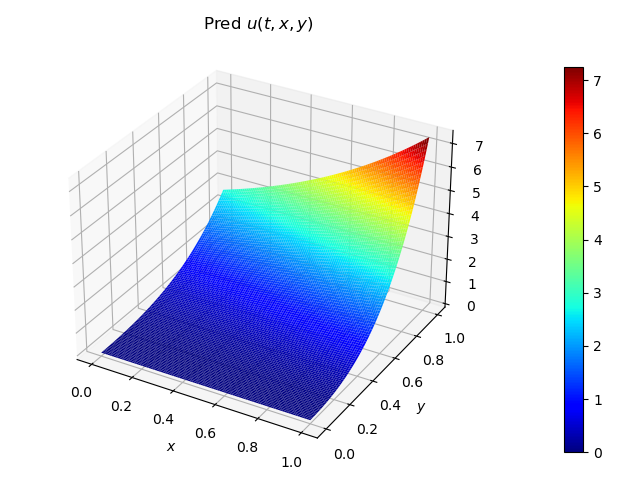}}
\subfigure[PMNN on $\mathrm{L} 2-1_\sigma$]{
\includegraphics[width=2.0in]{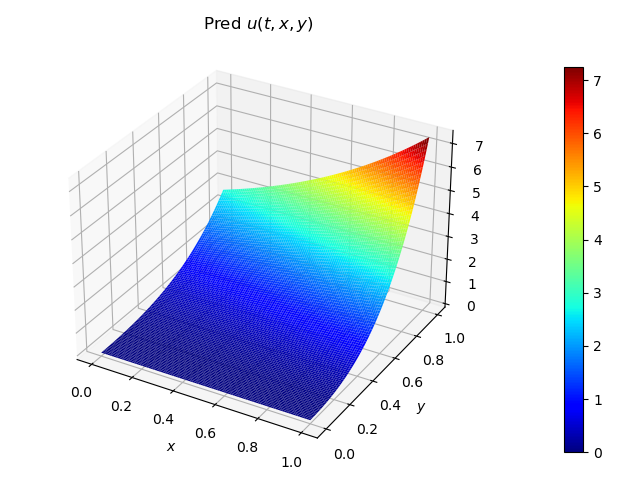}}
\caption{2D Single-term FPDE: the exact solution and the predict solution when $t=1.00$.} 
\label{2D_PDE_pred}
\end{figure}

Fig.\ref{2D_PDE_error} depicts the errors of the two PMNN models at $t=1$. It can be observed that the error of PMNN on $\mathrm{L} 1$ is relatively larger near the center of the spatial plane, while it decreases as it approaches the boundaries. On the other hand, the error of PMNN on $\mathrm{L} 2-1_\sigma$ exhibits less fluctuation.
\begin{figure}[!h]
\centering
\subfigure[PMNN on $L1$]{
\includegraphics[width=2.5in]{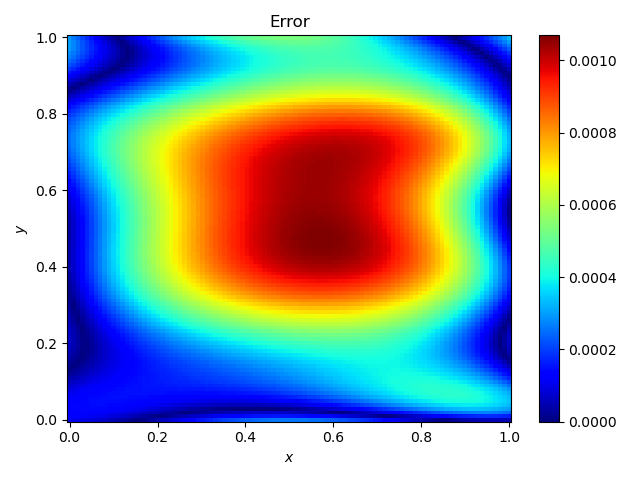}}
\subfigure[PMNN on $\mathrm{L} 2-1_\sigma$]{
\includegraphics[width=2.5in]{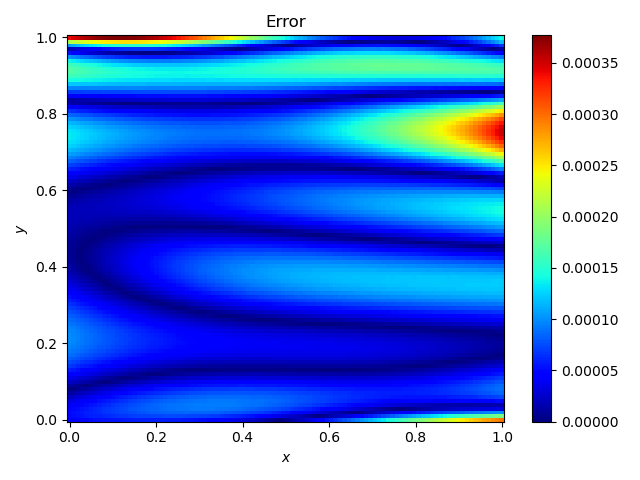}}
\caption{2D Single-term FPDE: the error presentation of PMNN when $t=1.00$.} 
\label{2D_PDE_error}
\end{figure}

Fig.\ref{2D_PDE_loss} illustrates the trends in the loss functions of the two models. The first two subfigures provide a detailed view of the changes in the individual loss components, while the last subfigure depicts the overall loss of the models. Consistent with the experimental results in Example \ref{example1}, it is evident that PMNN on $\mathrm{L} 2-1_\sigma$ exhibits a faster convergence speed.
\begin{figure}[!h]
\centering
\subfigure[PMNN on $L1$]{
\includegraphics[width=2.0in]{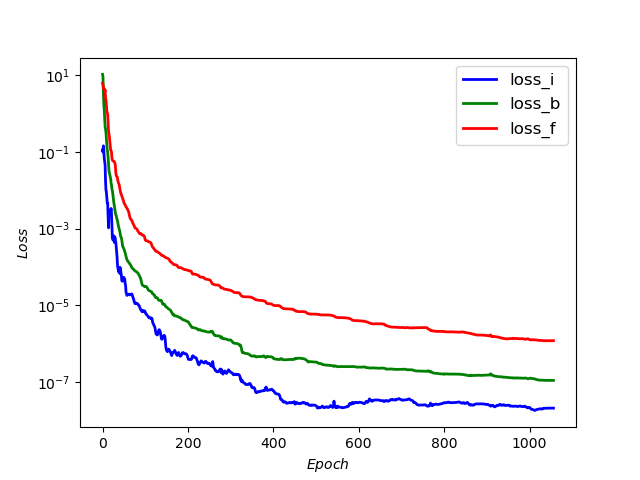}}
\subfigure[PMNN on $\mathrm{L} 2-1_\sigma$]{
\includegraphics[width=2.0in]{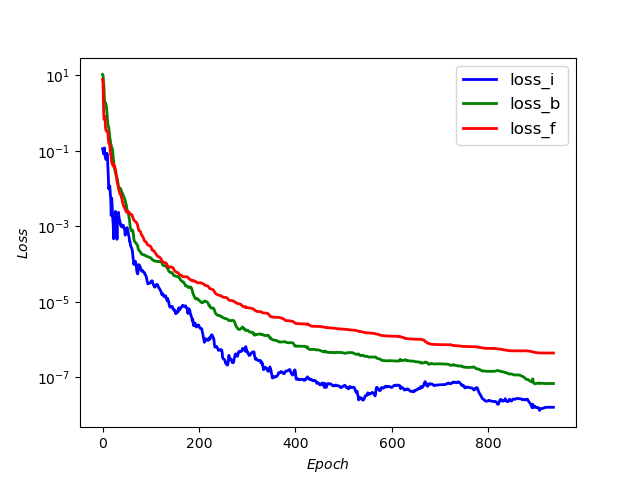}}
\subfigure[Comparison of the two models]{
\includegraphics[width=2.0in]{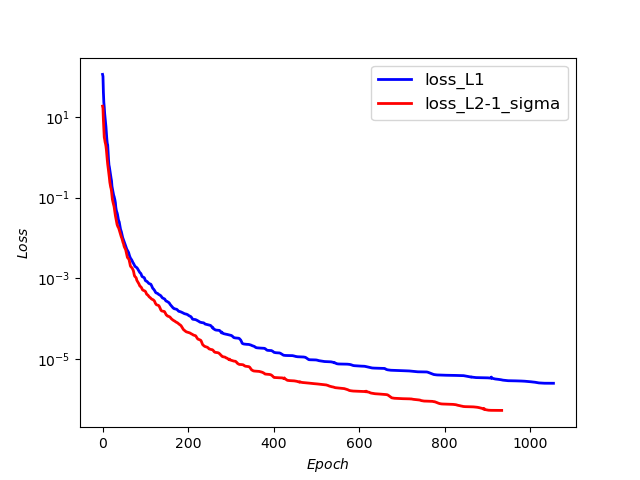}}
\caption{2D Single-term FPDE: the trend of the loss function with the number of iterations.} 
\label{2D_PDE_loss}
\end{figure}

Table \ref{tab:2D_PDE_time} presents the number of iterations and training time for both models when different orders of $\alpha$ are chosen. It can be observed that PMNN on $\mathrm{L} 1$ converges faster when $\alpha=0.25$, while PMNN on $\mathrm{L} 2-1_\sigma$ converges faster when $\alpha=0.5$ and $\alpha=0.75$. This indicates that the performance of the two models is influenced by the order of the fractional derivative in this example. This conclusion is consistent with Example \ref{example2}.
\begin{table}[!h] 
  \caption{The number of iterations and training time of two PMNN for 2D single-term FPDE}
  \label{tab:2D_PDE_time}
  \begin{center}
    \begin{tabular}{c cc cc}
    \toprule
    \multirow{2}{*}{}& 
    \multicolumn{2}{c}{Iter} & \multicolumn{2}{c}{Training Time(s)}\cr
    \cmidrule(lr){2-3} \cmidrule(lr){4-5} 
    & $L1$ & $L2-1_\sigma$ & $L1$ & $L2-1_\sigma$ \cr
    \midrule
    $\alpha=0.25$ & 1595 & 1875 & 287.62 & 341.82  \cr
    $\alpha=0.50$ & 1056 & 935 & 315.21 & 280.81  \cr
    $\alpha=0.75$ & 1076 & 723 & 347.18 & 218.5  \cr
    \bottomrule
    \end{tabular}
\end{center}
\end{table}

In the following, we fix $N_{t}=21$ and $\alpha=0.5$, and then vary the number of spatial nodes, $N_{x}$, to investigate their impact on the model. The corresponding $L^{2}$ relative errors are presented in Table \ref{tab:2D_PDE_x}. The analysis reveals that the model's performance is unaffected by the number of spatial training points. Consequently, it is viable to employ a minimal number of spatial nodes during training, resulting in reduced computational costs.
\begin{table}[!h] 
  \caption{The $L^{2}$ error of two PMNN for 2D single-term FPDE}
  \label{tab:2D_PDE_x}
  \begin{center}
    \begin{tabular}{c cc}
    \toprule
    \multirow{2}{*}{$N_{x}$}& 
    \multicolumn{2}{c}{$\alpha=0.5$} \cr
    \cmidrule(lr){2-3}
    & $L1$ & $L2-1_\sigma$ \cr
    \midrule
    6   & 3.42e-04 & 7.03e-05 \cr
    11  & 3.16e-04 & 4.67e-05 \cr
    21  & 3.07e-04 & 4.16e-05 \cr
    41  & 3.10e-04 & 4.25e-05 \cr
    81  & 3.09e-04 & 6.76e-05 \cr
    \bottomrule
    \end{tabular}
   \end{center}
\end{table}

Lastly, by fixing $N_{x}=11$, we perform experiments with different values of $\alpha$ and varying $N_{t}$. The $L^{2}$ relative errors are presented in Table \ref{tab:2D_PDE}. In contrast to the previous two experiments, the errors of both models do not consistently decrease as the number of time nodes increases. Instead, they exhibit minor fluctuations. Overall, the PMNN on $\mathrm{L} 2-1_\sigma$ model demonstrates superior performance compared to the PMNN on $\mathrm{L} 1$ model.
\begin{table}[!h] 
  \caption{The $L^{2}$ error of two PMNN for 2D single-term FPDE}
  \label{tab:2D_PDE}
  \begin{center}
    \begin{tabular}{c cc cc cc}
    \toprule
    \multirow{2}{*}{$N_{t}$}& 
    \multicolumn{2}{c}{$\alpha=0.25$} & \multicolumn{2}{c}{$\alpha=0.50$} & \multicolumn{2}{c}{$\alpha=0.75$} \cr
    \cmidrule(lr){2-3} \cmidrule(lr){4-5} \cmidrule(lr){6-7}
    & $L1$ & $L2-1_\sigma$ & $L1$ & $L2-1_\sigma$ & $L1$ & $L2-1_\sigma$\cr
    \midrule
    11  & 2.50e-04 & 6.23e-05 & 8.13e-04 & 6.15e-05 & 2.34e-03 & 5.00e-05 \cr
    21  & 8.27e-05 & 4.01e-05 & 3.16e-04 & 4.67e-05 & 1.01e-03 & 4.17e-05 \cr
    41  & 4.07e-05 & 1.66e-04 & 1.21e-04 & 3.91e-05 & 4.32e-04 & 4.77e-05 \cr
    81  & 2.34e-05 & 3.32e-04 & 5.26e-05 & 5.25e-05 & 1.76e-04 & 5.08e-05 \cr
    101 & 3.04e-05 & 5.61e-05 & 4.56e-05 & 5.62e-05 & 1.35e-04 & 5.68e-05 \cr
    \bottomrule
    \end{tabular}
   \end{center}
\end{table}

}
\end{example}

\section{Conclusions}
\label{4}
In this paper, we introduce PMNN, an iteration scheme approximation method based on physical model-driven neural networks, to solve FDEs. This iteration scheme leverages the physical information embedded in the equations, enabling the problem of solving FDEs to be reframed as a learning task for the PMNN model. Specifically tailored for Caputo FDEs, PMNN overcomes the limitations of automatic differentiation techniques in neural networks for solving fractional derivatives. It combines the efficiency of traditional interpolation approximation and harnesses the powerful fitting capabilities of neural networks. Through three numerical experiments, we demonstrate the excellent performance of the proposed model. Moreover, we present two variations of the model, and the numerical experiments show their distinct merits in various scenarios. Hence, in practical applications, the selection of the suitable model can be selected to cater to specific requirements. In spite of that, PMNN on $\mathrm{L} 2-1_\sigma$ exhibits particularly appealing performance in the majority of cases.

PMNN perfectly merges physical model-driven neural networks with interpolation approximation techniques for fractional derivatives, offering a novel approach for numerically solving FDEs. The proposed model currently applies only to the solution of single-term temporal fractional differential equations. Furthermore, some interesting phenomena observed in the experimental results still lack a clear explanation, presenting unresolved challenges for future investigation. In the future, investigating neural network-based physical model-driven methods for multi-term FDEs and spatial fractional-order differential equations could provide a potential direction for further research.

\section*{CRediT authorship contribution statement}
$\textbf{Zhiying Ma}:$ Methodology, Software, Writing-original draft. $\textbf{Jie Hou}:$ Methodology, Writing – review \text{\&} editing. $\textbf{Wenhao Zhu}:$ Supervision, Writing-review \text{\&} editing. $\textbf{Yaxin Peng}:$ Supervision, Writing-review \text{\&} editing. $\textbf{Ying Li}:$ Conceptualization, Funding acquisition, Writing-review \text{\&} editing.
\section*{Acknowledgments}
This work is supported by the National Key Research and Development Program of China (No.2021YFA1003004).

\section*{Declaration of competing interest}
The authors declare that they have no known competing financial interests or personal relationships that could have appeared to influence the work reported in this paper.

\bibliographystyle{ieeetr}
\bibliography{reference-final}
\end{document}